\documentclass[nohyperref]{article}

\usepackage{microtype}
\usepackage{graphicx}
\usepackage{subfigure}
\usepackage{booktabs} 
\usepackage{enumitem}   
\usepackage{float}
\usepackage{hyperref}

\usepackage{makecell}

\usepackage[accepted]{icml2023}

\usepackage{amsmath}
\usepackage{amssymb}
\usepackage{mathtools}
\usepackage{amsthm}
\usepackage{multirow}

\usepackage[capitalize,noabbrev]{cleveref}

\DeclareUnicodeCharacter{2212}{-}

\newcommand{\dgsr}{\texttt{DGSR}}
\newcommand{\ours}{\texttt{DGSR-MCTS}}

\usepackage[ruled,vlined,noend]{algorithm2e}

\usepackage{xcolor}
\definecolor{C0}{HTML}{1f77b4}
\definecolor{C1}{HTML}{ff7f0e}
\definecolor{C2}{HTML}{2ca02c}
\definecolor{C3}{HTML}{d62728}
\definecolor{C4}{HTML}{9467bd}
\definecolor{C5}{HTML}{8c564b}
\definecolor{C6}{HTML}{e377c2}
\definecolor{C7}{HTML}{7f7f7f}
\definecolor{C8}{HTML}{bcbd22}
\definecolor{C9}{HTML}{17becf}

\usepackage[normalem]{ulem}

\theoremstyle{plain}

\theoremstyle{definition}

\theoremstyle{remark}

\usepackage{pdfpages}

\usepackage[textsize=tiny]{todonotes}

\icmltitlerunning{Deep Generative Symbolic Regression with MCTS}

\begin{document}

\twocolumn[
\icmltitle{Deep Generative Symbolic Regression with Monte-Carlo-Tree-Search}

\icmlsetsymbol{equal}{*}

\begin{icmlauthorlist}
\icmlauthor{Pierre-Alexandre Kamienny}{meta,isir}
\icmlauthor{Guillaume Lample}{meta}
\icmlauthor{Sylvain Lamprier}{angers}
\icmlauthor{Marco Virgolin}{cwi}

\end{icmlauthorlist}

\icmlaffiliation{meta}{Meta AI, Paris, France}
\icmlaffiliation{isir}{ISIR MLIA, Sorbonne Université, France}
\icmlaffiliation{angers}{LERIA, Université d’Angers, France}
\icmlaffiliation{cwi}{Centrum Wiskunde \& Informatica, the Netherlands}

\icmlcorrespondingauthor{Pierre-Alexandre Kamienny}{pakamienny@meta.com}

\icmlkeywords{Machine Learning, ICML}

\vskip 0.3in
]



\printAffiliationsAndNotice{} 

\begin{abstract}

Symbolic regression (SR) is the problem of learning a symbolic expression from numerical data.
Recently, deep neural models trained on procedurally-generated synthetic datasets showed competitive performance compared to more classical Genetic Programming (GP) algorithms. Unlike their GP counterparts, these neural approaches are trained to generate expressions from datasets given as context. 
This allows them to produce accurate expressions in a single forward pass at test time.
However, they usually do not benefit from search abilities, which result in low performance compared to GP on out-of-distribution datasets.
In this paper, we propose a novel method which provides the best of both worlds, based on a Monte-Carlo Tree Search procedure using a context-aware neural mutation model, which is initially pre-trained to learn promising mutations, and further refined from successful experiences in an online fashion.
The approach demonstrates state-of-the-art performance on the well-known \texttt{SRBench} benchmark. 

\end{abstract}

\section{Introduction}
Symbolic Regression (SR) is the problem of simultaneously finding the structure 
(operators, variables) and optimising the parameters (constants) of an expression that describes experimental data in an interpretable manner. 
SR can produce human-readable expressions that are particularly useful in natural sciences, e.g., materials sciences \cite{wang2019symbolic,kabliman2021application,ma2022evolving} or physics \cite{schmidt2009distilling,vaddireddy2020feature,sun2022symbolic,cranmer2020discovering,hernandez2019fast,udrescu2020ai}. 
The recent benchmarking effort \texttt{SRBench} \cite{la2021contemporary} has shown that SR algorithms can additionally outperform over-parametrized models, e.g., decision-tree ensembles or neural networks, on a set of real-world and synthetic datasets.

SR is a challenging task, which implies composing inherently discrete objects (operators and variables) to make the resulting expression fit well the given data.  
As seeking the optimal composition is intractable~\cite{virgolin2022symbolic}, typical approaches to SR are based on heuristics.
In fact, the leading algorithms on \texttt{SRBench} are modern versions of genetic programming (GP)~\cite{koza1994genetic}, a class of genetic algorithms where evolving solutions need to be executed to determine their quality (i.e., SR expressions need to be evaluated on the data to determine their quality of fit).

Lately, there has been a growing interest in the SR community for neural network-based approaches.
For example, \citet{udrescu2020ai} use neural networks (NNs) to explicitly detect data properties (e.g., symmetries in the data) useful to reduce the search space of SR.
Other approaches, which we group together under the name of Deep Generative Symbolic Regression (\dgsr), have used NNs to learn to generate expressions directly from the data. 
Similarly to GP, DSR~\cite{petersen2019deep} and \cite{mundhenk2021symbolic} faces a {\em tabula rasa} setup of the problem for each new dataset. 
On the other hand, 
several  \dgsr{} approaches are inductive: 
they are pre-trained to predict an expression in a single forward pass for any new dataset, by feeding the dataset 
as input tokens \cite{biggio2021neural,valipour2021symbolicgpt,kamienny2022end}.
As such, these approaches have the appeal of generating expressions extremely quickly. However, their lack of a search component makes them unable to \emph{improve} for the specific dataset at hand.
This aspect can be particularly problematic when the given data is out-of-distribution compared to the synthetic data the NN was pre-trained upon. 
A promising direction to cope with the limitations of inductive \dgsr{} is therefore to include a suitable search strategy. 
The use of neural policies in Monte-Carlo Tree Search (MCTS) has led to improved exploration via long-term planning in the context of automated theorem proving with formal systems \cite{polu2020generative, lample2022hypertree}, algorithm discovery \cite{fawzi2022discovering}, and also games \cite{silver2016mastering,silver2018general}.
In the context SR, 
search tree nodes are expressions (e.g., $x_1 + x_2$), and edges between them represent mutations (e.g., $x_2 \rightarrow \left(7 \times x_3\right)$, leading to the expression $x_1 + 7 \times x_3$).
\citet{white2015programming,sun2022symbolic} proposed a classic formulation of MCTS for SR, where possible mutations are pre-defined along with prioritization rules to decide which nodes to mutate in priority.
\citet{li2019neural} use a recurrent neural network to produce possible mutations, which is conditioned on shape constraints for the expressions, but not on the dataset.
Most similar to our approach is the study of \citet{lu2021incorporating}, that uses a pre-trained model to sample promising but rather simple mutations (up to one expression term).
However, the model is not fine-tuned on the specific dataset at hand as the search progresses. In our proposal, MCTS is augmented with neural policies that are both pre-trained and fine-tuned. Existing works have only showed good performance on simple benchmark problems (e.g., no observed competitive performance on \texttt{SRBench} real-world problems). Furthermore, we found in preliminary experiments that the combination of NN within MCTS with pre-training is key to achieve good performance.

\paragraph{Summary of the approach.} In this paper, we seek to overcome the limitations of \dgsr{}, by proposing a synergistic combination, which we call \ours{}, where MCTS is seeded with pre-trained \dgsr{}, and \dgsr{} is fine-tuned over time on multiple datasets simultanously as the search progresses.
A \textit{mutation} policy is responsible for producing mutations that expand expressions from the tree search. 
A \textit{selection} policy prioritizes which expression to mutate next, by trading-off between exploration (low number of visits) and exploitation (most promising expressions) using a critic network. 
The mutation policy first undergoes a pre-training phase. 
Then, both the mutation policy and the critic network are updated in online fashion, by leveraging results from search trials on new provided datasets.

We first position our approach in an unifying view of SR, that encapsulates both earlier work in GP and more recent deep learning-based approaches.
Then, we describe our approach in detail, and evaluate it on the \texttt{SRBench} benchmark.
We show that our approach achieves state-of-the-art performance, and discovers expressions that are both simple and accurate.

\section{Unifying View of of SR}
\label{sec:background}
\looseness=-1 

\begin{table*}[ht!]
    \centering
    \caption{ Unifying view of SR.  $\theta$ 
    represents 
    weights of a probabilistic  neural network that embodies $P_f$. $\psi$ is parameters of a critic network (see \Cref{sec:method}).  
 }
    \resizebox{\textwidth}{!}{%
    \begin{tabular}{c|c|c|c}
    \toprule
    Algorithm &  Pre-train of $P^0_f$ & $P^t_f$ conditioned by   & Update of $P^t_f$ \\
    \midrule
    GP \cite{poli2008field} & No & Population \&  Genetic operators & Selection operators \\\hline
    EDA \cite{kim2014probabilistic} & No & Explicit Factorization & Selection operators \\\hline
    \texttt{E2E} \cite{kamienny2022end} & SSL on synthetic data & $\mathcal{D}$ \& $\theta^0$ & No  \\\hline
    DSR\cite{petersen2019deep} & No & $\theta^t$ & Update $\theta^t$ with policy gradients\\\hline
    uDSR \cite{landajuelaunified} & SSL on synthetic data & $\mathcal{D}$ \& $\theta^t$ & Update $\theta^t$ with policy gradients \\\hline
    \multirow{ 2}{*}{\ours{} (Ours)} & SSL and MCTS & MCTS  & Update $\theta^t$ \& $\psi^t$ \\ 
    & on synthetic data & using $\mathcal{D}$ \& $\theta^t$ \& $\psi^t$ 
    & via Selection \& Imitation \\
    \bottomrule
    \end{tabular}
    }
    \label{tab:unifying view}
\end{table*}

Let us denote by ${\cal F}$ the family of symbolic expressions\footnote{Note that even though different expressions may be functionally equivalent, this is normally not taken into account in existing approaches, as determining functional equivalence can be undecidable~\cite{buchberger1982algebraic}.}  that form the search space of the SR problem. Generally speaking, ${\cal F}$ is defined by the set of building blocks from which expressions can be composed \cite{virgolin2022symbolic}. These usually include constants (possibly sampled from a distribution, e.g., $\mathcal{N}(0,1)$), variables, and basic operators as well as trigonometric or transcendental ones (e.g., $+,-,\times,\div,\sin,\cos,\exp,\log$).

A common formulation of SR poses to seek a well-fitting expression for the dataset at hand.
We now generalize this idea to the case where good performance is sought across multiple datasets (as one usually seeks a generally-competent search algorithm rather than a dataset-specific one).
Given $\Omega$ a distribution over datasets $\mathcal{D}=\left\{(\mathbf{x}_i, y_i)\right\}_{i\leq N}$, with $(\mathbf{x_i}, y_i)\in \mathbb{R}^d \times \mathbb{R}$ a given point  (descriptor, image) from a specific dataset $\mathcal{D}$, and a limited budget for the exploration process $T$, the general objective of SR is to define an algorithm that produce distributions of expressions $f \in {\cal F}$, that minimize the following theoretical risk at step $T$: 
\begin{equation}
 {\cal R}_{\Omega, {\cal F}}=\mathbb{E}_{{\cal D} \sim \Omega({\cal D})} \mathbb{E}_{f \sim P^T_f({\cal D})}[{\cal L}(f,{\cal D})] 
\label{eq:sr_objective}
\end{equation}
with $P^T_f({\cal D})$ the final distribution over expressions provided by the considered algorithm.
A typical loss function $\mathcal{L}$ considered in the SR literature is the negative $R^2$: 
\begin{equation}
    R^2(f,{\cal D})=1-\frac{\textit{MSE}(\mathbf{y}, f(\mathbf{x}) )}{\textit{VAR}(\mathbf{y})} = 1-\frac{\sum_i \left( y_i-f(\mathbf{x}_i) \right)^2}{\sum_i (y_i-\overline{y}_i)^2}.
\label{eq:r2_definition}
\end{equation} 
where $\overline{y}_i=(1/N)\sum_{i\leq N} y_i$. 

Considering algorithms that start from $P^0_f({\cal D})$ to incrementally build $P^T_f({\cal D})$, the problem can be decomposed into two main steps: 1) define an accurate starting point $P^0_f({\cal D})$ for the search process; 
2) specify the exploration process that allows to update $P^0_f({\cal D})$ to form $P^T_f({\cal D})$ in a maximum of $T$ iterations (search trials).
Some approaches in the literature only consider the first step (i.e., $P^T_f({\cal D})=P^0_f({\cal D})$ ---no search process).
Others only investigate step 2 (i.e., $P^0_f({\cal D})=P^0_f(\emptyset)$ ---no inductive context), as described in the following.

There are many ways to define $P^0_f$. 
For example, traditionally in GP, the dataset does not play a role (i.e., it is $P^0_f(\emptyset)$), but there exist different strategies to randomly initialize the population~\cite{looks2007behavioral,pawlak2016semantic,ahmad2018comparison}.
Since we deal with \dgsr{}, we focus on $P^0_f({\cal D})$ that entail pre-training from here on.
Note that $\Omega$ is unknown during this pre-training, and that we only observe its datasets at search time (datasets from \texttt{SRBench} in our experiments).

\subsection{Pre-training: How to define $P^0_f({\cal D})$}

The formulation of \cref{eq:sr_objective} is reminiscent of meta-learning or few-shot learning where the goal is to train a model on a set of training tasks such that it can solve new ones more efficiently \cite{schmidhuber:1987:srl,thrun2012learning,finn2017model}. In that vein, many approaches focus on learning a generative model to induce an implicit form of $P^0_f$. However, since $\Omega$ is unknown during pre-training, those models are usually trained on large synthetic datasets built from randomly generated expressions \cite{lample2019deep}.

Rather than focusing on the accuracy of the sampled expressions regarding some criteria such as \Cref{eq:r2_definition}, the vast majority of these approaches are neural language models
(and thus belong to the \dgsr{} family)
trained to express $P^0_f$ by auto-regressively predicting  a given ground truth expression from the training data (e.g., using a recurrent neural network or a transformer).

Although these methods tend to produce expressions similar to the synthetic ones seen during pre-training, they have the advantage of explicitly considering the dataset as an input: $P_f$ is conditioned on $\mathcal{D}$~\cite{biggio2021neural,valipour2021symbolicgpt,kamienny2022end}.
Consequently, pre-training approaches for \dgsr{} can produce expressions as a one-shot inference process by the simple action of sampling, eliminating the need for search iterations. 
However, sampling from $P^0_f$ is limited as the accuracy improvement stops after a small number of samples ($50$ in~\cite{kamienny2022end}), and can perform badly on out-of-domain datasets given at test-time. 

\subsection{Search Process: How to build $P^T_f({\cal D})$}

Given a dataset ${\cal D}$, the aim of any process at search time is to define a procedure to build an accurate distribution $P^T_f({\cal D})$ from the starting one $P^0_f({\cal D})$, via a mix of exploration and exploitation. 
Most approaches seek the maximization of $\mathbb{E}_{f \sim P^T_f}[R^2(f,{\cal D})]$.  
Several approaches additionally include in that objective a functional $C(f)$ that penalizes overly-complex  expressions~\cite{vladislavleva2008order,chen2018structural}. 
Even though there exists different definition of \textit{complexity} \cite{kommenda2015complexity,virgolin2020learning}, we consider  \textit{expression size} (i.e., the number of operators, variables and constants in the expression) as complexity measure. 

In that aim, the learning process of different SR algorithms  can be unified into a general, iterative framework: until the termination condition is not met (e.g., runtime budged exceeded or satisfactory accuracy reached), 
\begin{enumerate}[label=(\roman*)]
\item sample one or more symbolic expressions $f \in \mathcal{F}$ using the  current probability distribution $P^t_f$ at step $t$;
\item update $P^t_f$ using statistics of the sample, such as the accuracy of the expression.
\end{enumerate}
Steps (i) and (ii) constitute an iteration of the framework. 
Different SR algorithms have considered various definitions of $P^t_f$ and strategies to update it. Its definition can be implicit, explicit, or learned, and be biased towards certain expression structures. 
Table \ref{tab:unifying view} provides a non-exhaustive summary of popular algorithms, which we elaborate upon in the following paragraphs.

Genetic Programming (GP) is a meta-heuristic inspired by natural evolution and is a popular approach to SR~\cite{koza1994genetic,poli2008field}.
GP implements step (i) by 
maintaining a population of expressions which undergoes stochastic modifications (via crossover and mutation), and step (ii) by selection, i.e., stochastic survival of the fittest, where better expressions are duplicated and worse ones are removed.
In other words, sampling and updating $P^t_f$ is implicitly defined by the heuristics that are chosen to realize the evolution.


Estimation of Distribution Algorithms (EDAs) attempt to be more principled than GP by explicitly defining the form of $P^t_f$ ~\cite{salustowicz1997probabilistic,hemberg2012investigation,kim2014probabilistic}. 
Normally, $P^t_f$ is factorized according to the preference of the practitioner, e.g., expression terms can be modelled and sampled independently or jointly in tuples, or chosen among multiple options as the search progresses, using, e.g., the minimum description length principle~\cite{harik1999linkage}.
EDAs use 
methods similar to those of GP to realize step (ii).



\looseness=-1 
DSR \cite{petersen2019deep} proposed an approach where $P^t_f$ is realized by an NN, whose parameters $\theta$ are updated using policy gradients with the accuracy of sampled expressions as rewards. 
Though being very general and only biased by the model parametrization $\theta$, this approach was found to generate very short expressions with low accuracy on \texttt{SRBench} (see \cref{sec:experiments}). 
This is likely due to sparse reward and credit assignment issues typical of reinforcement learning \cite{sutton1984temporal}. \cite{mundhenk2021symbolic} adds GP search samples on top of DSR.
Very recently, the work by \citet{petersen2019deep} was integrated with a pre-training component~\cite{landajuelaunified}, albeit in a large pipeline that also includes GP, NNs for search space reduction~\cite{udrescu2020ai}, and linear regression.

In the recent benchmarking effort \texttt{SRBench} by~\citet{la2021contemporary}, modern GP algorithms have shown to be the most successful.
An interesting property that distinguishes GP from other \dgsr{} approaches is that its crossover and mutation operators tend to edit existing expressions (thus preserving substantial parts of them), rather than sampling them from scratch~\cite{koza1994genetic,poli2008field}.
The algorithm we develop in \Cref{sec:method} expands expressions over time, similarly to GP.
At the same time, our algorithm is powered by pre-training and its ability to learn over time, a feature missing from GP and other approaches.

\section{Method}
\label{sec:method}
\looseness=-1 Following the unified framework of SR detailed in previous section, we derive our method, \ours{}, as an expert-iteration algorithm \cite{anthony2017thinking}, which iteratively 1) samples expressions from $P^t_f$ 
and 2) updates $P_f^t$  by improving the distribution via imitation learning (i.e., log-likelihood maximization of solution expressions). 

Recall that in \dgsr{} methods, sampling expressions from $P^t_f$ involve producing tokens step-by-step by sampling from a next-token distribution with techniques such as Monte-Carlo sampling or Beam-search. Turning pre-trained \dgsr{} methods into ones that can update $P^t_f$ with expert-iteration is a challenging task as 1) reward signal can be very sparse (i.e., very few sequences may correspond to accurate expressions for the given dataset); 2) 
such left-to-right blind way of decoding does not allow for accurate planning; 3) using accuracy objectives to guide the decoding would be difficult with such auto-regressive approaches, since intermediate sequences of tokens are not valid expressions in that setting.  


Rather, we propose to derive $P^t_f$ as the distribution induced by a Monte-Carlo Tree Search (MTCS) process \cite{browne2012survey}, which we iteratively refine via imitation learning  on samples it produces that solve the problem. 
Our MCTS process also considers expression mutations, following a mutation policy  $M_\theta^t$ which deals with transformations of existing expressions, rather than a greedy concatenation of tokens from a given vocabulary, allowing a more systematic evaluation of intermediate solutions for a more guided exploration. This section first explains our MCTS search process then the way we pre-train the mutation policy from synthetic data.


\begin{figure*}[htbp]
  \centering
 \includegraphics[width=2.0\columnwidth]{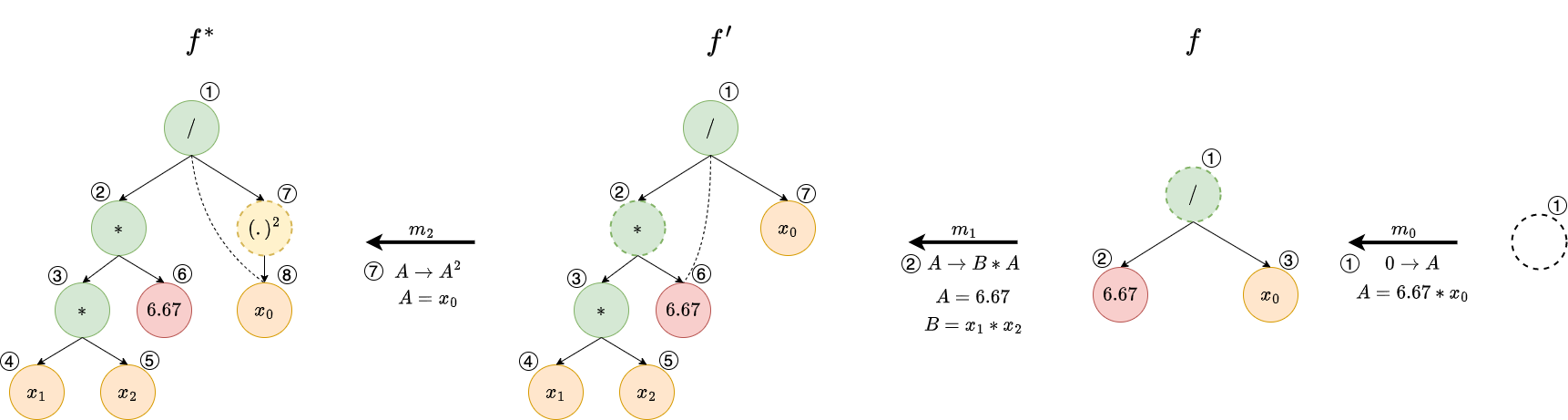}
    \caption{
 Example of data generation to train the mutation model.
  Given a starting ground-truth expression (e.g., $f^*(x_0,x_1,x_2)=6.67 x_1 x_2/x_0^2$ as a tree, we procedurally dismantle the tree until no node is left.
  This is done by, at each step (red arrows), a) picking a node (dashed contour), b) removing the picked node and, if the operator is binary, additionally remove the subtree rooted in one of the two child nodes $B$, 
  c) adding an edge (black dotted line) between the parent node and the remaining child node $A$ to obtain a well-formed expression.
  When the picked node is the root node, the entire tree is removed, and the dismantling stops.
  Then, we train the mutation model to assemble the tree back via subsequent mutations (green arrows), which revert the dismantling process.
  The mutation model is conditioned on the current tree (initially empty) as well as the dataset $\mathcal{D}$.
}
  \label{fig:data_generation}
\end{figure*}

\subsection{MCTS Sampling} 
\label{subsec:sr_as_tree}

Our sampling process of new expressions is derived from an MCTS process,  where each node of the search tree\footnote{Not to be confused with tree-based representations of expressions, which we discussed in the previous section.} corresponds to an expression $f \in {\cal F}$. Following a similar process as in \cite{lample2022hypertree}, our MCTS process is made of three steps: 1) Selection, 2) Expansion, 3) Back-propagation, that we detail below. 
In our MCTS, $M_\theta$ is used to sample promising mutations that populate the MCTS search tree via expansions (described below). Besides $M_\theta$, we additionally leverage a critic network $C_\psi$ to assess how likely an expression can lead to a solution.
We use the same NN (weights are shared) to realize $M_\theta$ and $C_\theta$, with the exception that $C_\theta$ includes an additional head that is trained during the process to output a value in $\mathbb{R}$.
These three steps are iteratively repeated $1000$ times, a period\footnote{This corresponds to a single step i) in the unifying framework of \cref{sec:background}.} that we call \textit{search trial}, before $M_\theta$ and $C_\psi$ is updated.

\paragraph{Selection} The selection step of MCTS aims at following a path in the search tree, from its root, i.e. the empty expression, to a leaf in the search tree, that trades off between exploration and exploitation. Following PUCT \cite{silver2016mastering}, at each new node $f$, we select its child $f'$ that maximizes: 
\begin{equation}
 V(f') + p_{uct} E(f') M_\theta(f'|f, \mathcal{D}, \theta)
\label{eq:mcts_policy}
\end{equation}
with $E(f')=\frac{\sqrt{\sum_{f'\in \text{child}(f)}N(f')}}{1 + N(f')}$. 
$p_{uct} \in \mathrm{R}^+$ is the exploration coefficient, $N(f)$ is the number of times $f$ was selected so far during the search, and $V(f)$ is an estimate of the expected value of the expression, expressed as $v(f)/N(f)$, with $v(f)$ accumulating values of all the nodes depending on $f$ in the tree search. 
This selection criteria balances exploration and exploitation, controlled by hyper-parameter $p_{uct}$.  

\paragraph{Expansion} Once the selection step reaches a leaf $f$ of the tree, this leaf is expanded to produce new child nodes by applying $K$ mutations to $f$,  sampled from $M_\theta({\cal D},f, \theta)$. This leads to modified expressions $\{f'_k\}_{k \leq K}$. In our case, the distribution $P_f(f'|f, \mathcal{D}, \theta)$ is induced by the application of $m \sim M_\theta$ on $f$, resulting in $f'=m(f)$. For each $k \leq K$, we add a node $f'_k$ to the search tree, as well as an edge between $f'_k$ and $f$, labeled $m_k$. 

\vspace{5pt}
Each expression resulting from an expansion is evaluated (with or without constant optimization as discussed in \Cref{subsec:indomain}) to check whether it \emph{solves} the SR problem. Here, we assess whether a relatively high accuracy is reached to determine whether the expression is a solution for the given dataset (we use $R^2 \geq 0.99$ in our experiments).
Nodes that solve the problem obtain a value $v(f) = 1$. Others obtain an estimate from the critic network: $v(f)=C_\psi({\cal D},f)$. 
We remark that a simpler strategy is to define $v(f)$ as the accuracy of $f$, i.e. $v(f) = R^2(f, \mathcal{D})$. However, this strategy usually induces deceptive rewards, because a few mutations that lead to less accurate expressions (e.g., akin to sacrificing pieces in chess) may be needed before a very accurate expression is found (resp., we obtain a winning position). We confirmed that our strategy outperforms the naive use of accuracy in preliminary experiments

\paragraph{Back-Propagation} After each expansion, the value of every ancestor $f'$ of any new node $f$ is updated as $V(f')\leftarrow V(f') + v(f)$.
Note that this form of back-propagation regards weighing the nodes of the MCTS search tree, and should not be confused with the algorithm used to train NNs.


\vspace{5pt}
As mentioned before, selection, expansion, and back-propagation are repeated $1000$ times, after which the search trial is completed. At completion, the parameters of $M_\theta$ and $C_\psi$ are updated as described in \cref{subsec:learning_critic}. Finally, the MCTS search tree is reset, and built anew using updated $M_\theta$ and $C_\psi$ during the next trial.

\subsection{Learning critic $C_\psi$ and $M_\theta$}
\label{subsec:learning_critic}
After each search trial, we update the two parametrized components $C_\psi$ and $M_\theta$. 
To that end, training samples from the previous search trials are stored in two separate first-in-first-out queues: a buffer stores mutation sequences $(f^{(\tau)}, m_{t})$ that produced a solution expression $f^*$ to update $M_\theta$\footnote{If multiple such sequences exist, we select the one with the smallest number of mutations.}, the other contains $V$ values of nodes. 
For the latter, nodes that lead to a solution expression $f^*$ are assigned a score of $1.0$. 
Others are considered for training only if their visit count is higher than a given threshold $N^{min}_\text{visits}$, in order to focus training on sufficiently reliable estimates. At each training step, batches containing an equal proportion of mutation and critic data points are sampled from the queues to train $M_\theta$ and $C_\psi$ respectively. Both training objectives are weighted equally. To prevent any mode collapse, we continue sampling training examples from the supervised data used to pre-train the mutation model $M_\theta$, as described in \cref{subsec:method_mutation}.



Note that even though the pre-training data generation is biased in different ways, stochasticity of $M_\theta$ enables its adaptation over search trials thanks to its updates; for instance, even if $M_\theta$ was trained to output mutations with arguments of size $B \approx 10$ can learn mutations of size $1$, and vice-versa. 
As a result,  $M_\theta$ can automatically learn the appropriate (and dataset-dependent) size of mutations through MCTS search.

We set up our MCTS search to simultaneously operate on multiple datasets at the same time, so that $M_\theta$ and $C_\psi$ can leverage potential information that is shared across different SR instances (transfer learning).
Given a set of datasets, a \textit{controller} schedules a queue of datasets that includes a proportion of unsupervised (i.e. the ground-truth expression is not known) datasets to send to \textit{workers} in charge of handling search trials. 
To avoid catastrophic forgetting, we also continue training on pre-training examples from synthetic datasets i.e. with example mutations as described in \cref{subsec:method_mutation}, in the spirit of AlphaStar~\cite{vinyals2019grandmaster} or HTPS~\cite{lample2022hypertree} with human annotated data. $M_\theta$ and $C_\psi$ updates are tackled by \textit{trainers}.

    
\subsection{Mutation Policy} 
\label{subsec:method_mutation}
Our search process relies on a mutation policy $M^t_\theta$, a distribution over promising mutations of a given expression $f \in {\cal F}$. In what follows, we drop the $t$ index from $M_\theta$ for clarity. Expressions are represented as a tree structure and mutations act by modifying it. 

\paragraph{Definition of $M_\theta$.}

We define mutations as transformations that can be applied on any node of a given expression tree. 
\Cref{tab:operators} contains the list of considered 
transformations, where $A$ stands for an existing node of the expression tree and $B$ stands for a new (sub-)expression to be included in the tree. 
The $\xrightarrow[]{}$ symbol represents the replacement operation, e.g., $A \xrightarrow[]{} A + B$ means that node A from $f$ is replaced by a new addition node with $A$ as its left child and $B$ its the right one. Thus, a valid mutation from $M_\theta$ is a triplet $<A,op,B>$, where $A$ is a node from the expression tree, $op$ is an operation from \cref{tab:operators} and $B$ is a new expression whose generation is described below. 
A constant optimization step, detailed in \Cref{ap:constant_optimization}, can be performed after the mutation to better fit the data; we explore in \Cref{sec:experiments} whether including constant optimization improves the performance. 
We call mutation \textit{size} the size of the expression $B$.  
\begin{table}[htb]
    \centering
    \caption{Set of operators that can be applied on a sub-expression $A$ of a function, with optional argument $B$ as a new sub-expression to include in the tree structure. 0 refers to the root node of the function.}
    \small
    \begin{tabular}{c|c}
    \toprule
        Unary & $A \xrightarrow[]{} \cos(A), A \xrightarrow[]{} \sin(A),  A \xrightarrow[]{} \tan(A)$  \\
         & $A \xrightarrow[]{} \exp(A), A \xrightarrow[]{} \log(A)$ \\ 
        & $A \xrightarrow[]{} A^{0.5}, A \xrightarrow[]{} A^{-1}, A \xrightarrow[]{} A^2$,  $0 \xrightarrow[]{} B $ \\
        \hline
        Binary & $A \xrightarrow[]{} A+B,  A \xrightarrow[]{} A-B$  \\
         & $A \xrightarrow[]{} A*B, A \xrightarrow[]{} A/B$  \\
         & $A \xrightarrow[]{} B+A,  A \xrightarrow[]{} B-A$  \\
         & $A \xrightarrow[]{} B*A, A \xrightarrow[]{} B/A$  \\
    \bottomrule
    \end{tabular}
     \label{tab:operators}
\end{table}    

The mutation policy $M_\theta$ provides a distribution over possible mutations. 
Rather than having $B$ be generated completely at random, we parameterize $M_\theta$ so that it is $M_\theta: \Omega \times \mathcal{F}$, i.e. the mutations conditions on the dataset $\mathcal{D}$ and on the current expression $f$.
The dependance from $\mathcal{D}$ is akin to the approach in inductive \dgsr{} works, while the dependence on $f$ is novel. Both are passed as inputs to $M_\theta$
as a sequence of tokens, $f$ by its prefix notation and $\mathcal{D}$ as in  \cite{kamienny2022end}. We use the transformer-based NN architecture from \citet{kamienny2022end} but task the model to decode token-by-token a sequence $\omega$ (flattened version of $<A,op,B>$) 
until a EOS token is reached. $A$ is represented in $\omega$ as the index of that node (i.e., $\in [\![1, n]\!]$ for an expression that contains $n$ nodes).
While this may allow to output an invalid mutation expression, this happens very rarely in practice as shown in \Cref{ap:exceptions}, thanks to an efficient pre-training of the policy (described below). 

\looseness=-1 We remark that our mutation distribution is different from those that are commonly used in GP, in that the latter are not conditioned on $\mathcal{D}$ nor parameters (i.e., they are not updated via learning), and they can also shrink the size of an expression or keep it as is, whereas our mutations strictly increase the expression.
Note that it is possible to consider mutations that remove and/or replace parts of the expression, but we left exploring this to future work.
We also restrict our mutation process to only generate expressions with less than $60$ operators and without nesting operations other than the basic arithmetic ones $(+,-,\times,\div)$.

\paragraph{Pre-training of $M_\theta$.} Since our mutation policy $M_\theta$ is expected to produce mutations for a given expression, and not the final expression directly (as it is the case in the majority of \dgsr{} approaches), it requires a specifically designed pre-training process. 
To that end, pre-training labeled examples (dataset \& ground-truth expression with up to $10$ features) are first sampled from a hand-crafted generator~\cite{lample2019deep} as done in most pre-training NSR approaches (c.f. \cref{ap:data_generation}). 
Next, given a ground-truth expression $f^*$, we extract a sequence of mutations $[m_l]_{\leq L}$ that iteratively map the empty expression $f^{(0)}$ to the final expression $f^*$.
As illustrated in \Cref{fig:data_generation}, starting from the ground-truth expressions $f^*$, we deconstruct $f^*$ by procedurally removing a node (and if the node is binary also one of its child subtree $B$) from the current $f$ until we get to the empty expression $f^{(0)}$. 
After reversing this sequence, we obtain a training sequence of expressions and mutations $f^{(0)}\xrightarrow[]{m_1}f^{(1)}\xrightarrow[]{m_2}f^{(2)}\xrightarrow[]{m_3}\ldots \xrightarrow[]{m_{L}}f^{(L)} = f^*$ (more details in \cref{ap:data_generation}).
After tokenization, every mutation $m_l$ 
serves as target for the pre-training process: $M_\theta$ is classically trained as a sequence-to-sequence encoder-decoder, using a cross-entropy loss, to sequentially output each token $w^e_l$ from $m_l$, given the considered dataset $\cal D$, the previous expression $f^{(l-1)}$, and the sequence of previous tokens $w^{(<e)}_l$ from the target operator $m_l$. 

\section{Experiments}
\label{sec:experiments}
In this section, we present the results of \ours{}. We begin by studying the performance on test synthetic datasets. Then, we 
present results 
on the \texttt{SRBench} datasets.


\subsection{Analysis on synthetic datasets}
\label{subsec:indomain} 
In this sub-section, we consider a set of $1000$ unseen synthetic expressions of which half are \textit{in-domain} (exactly same generator described in \cref{sec:method} and \cref{ap:data_generation}) and half are \textit{out-of-domain} (bigger expressions with up-to $40$ operators instead of $25$). 
We provide a set of explorative experiments to bring insights on how different hyper-parameters contribute to the performance, as well as to select a good configuration of hyper-parameters for evaluation on \texttt{SRBench}. 
In what follows, we always select the best expression on a given dataset by evaluating the accuracy of each expression on the training set; as mentioned before, we consider a dataset to be \textit{solved} if the $R^2$ achieved by the best expression is greater than $0.99$.
Pre-training was performed on 8 GPUs for a total time of 12 days. We controlled overfitting on the training set of expressions by i) using a sufficiently large training dataset, ii) controlling the cross-entropy loss and prediction accuracy on a held-out validation set of expressions.
Each run in this subsection was obtained by MTCS search trials (which fine-tune $M_\theta$ and $C_\psi$) with a time limit of 24 hours, using $4$ trainers (1 GPU/CPU each), $4$ MCTS workers (1 GPU/CPU each).

\paragraph{Breadth or depth?} First, we analyze whether, given a pre-trained $M_\theta$,  
it is more desirable to explore in breadth or in depth the search tree.  The number of samples  implicitly influences the breadth/depth trade-off; 
the larger the number of samples, the more it will be encouraged to explore, whereas when there is little number of samples $K$, it is forced to go deep. 
We run a single search trial of $2000$ iterations 
for different values of $K \in \{1,2,8,16,32\}$.

To compare these different configurations in a fair manner, we make a few design choices that we justify here. 
First, as shown in earlier work \cite{dick2020feature,kommenda2020parameter,kamienny2022end} and later in this section, optimization of constants contained in the expression  
can be important to reach  high accuracy levels in SR. The deeper in the search tree an expression is, the bigger the expression is, as well as the larger the number of constants it includes, which can add degrees of freedom to the optimization.
Because of this, depth can be expected to outperform breadth. 
For this reason and also because we consider constant optimization in a following ablation, we choose not to optimize constants in this experiment.
Secondly, we impose that each configuration considers the same number of expressions. 
We realize this by allowing only a subset of $K$ expressions to be visited out of the 32 that are sampled during expansion. As shown in \cref{tab:breadth_or_depth}, a choice of $K$ that is between $8$ and $16$  seems to be the best compromise in both in-domain and out-of-domain datasets, therefore we will sample $K \in [\![8, 16]\!]$ before each expansion in what follows.

\begin{table}[h]
    \caption{\textbf{Percentage of solved datasets for different $K$}}
    \centering
    \begin{tabular}{c| c | c}
    \toprule
    $K$ & In-Domain & Out-of-domain  \\
    \midrule
    1 (greedy) & $9.6$  & $0.8$ \\
    2 & $10.8$  & $2.4$ \\
    8 & $44.6$  & $\mathbf{19.6}$\\
    16  &$\mathbf{54.0}$ & $18.4$ \\
    32  &$42.8$& $10.2$ \\
    \bottomrule
    \end{tabular}
    \label{tab:breadth_or_depth}
\end{table}

\paragraph{Big or small mutation sizes?}
Secondly, we do ablations on three $M_\theta$ models pre-trained on mutation examples generated via different strategies with varying mutation sizes (as defined in \cref{subsec:method_mutation}); the higher in the expression tree a node is picked (as described in step a) of the caption of \Cref{fig:data_generation}), the bigger the mutation tends to be.
We consider \ours{}@$1$ (respectively \ours{}@$10$), a model pre-trained on mutation sizes $1$ (respectively approximately\footnote{Exact mutation size cannot be guaranteed without special expression tree structures.} $10$), and finally \ours{}@$\infty$, a model trained to output the target expression in a single iteration.
Note that \ours{}@$\infty$ is essentially reduces to approaches like those in~\cite{biggio2021neural,kamienny2022end}, as $M_\theta$ is tasked to predict the entire expression $f^*$ from scratch $f^{(0)}$, while updating $M_\theta$ with expert-iteration as described in \cref{sec:method}.

Interestingly, \cref{tab:model_sizes} shows mutation size of $10$ performs better than size $1$ both in-domain and out-of-domain and that the \ours{}@$\infty$ does not generalize to ouf-of-domain datasets, confirming the importance of search. 

\begin{figure*}[!ht]
\centering
\renewcommand{\arraystretch}{0}
\setlength{\tabcolsep}{0pt}
\begin{tabular}{cc}
 \includegraphics[width=0.99\columnwidth]{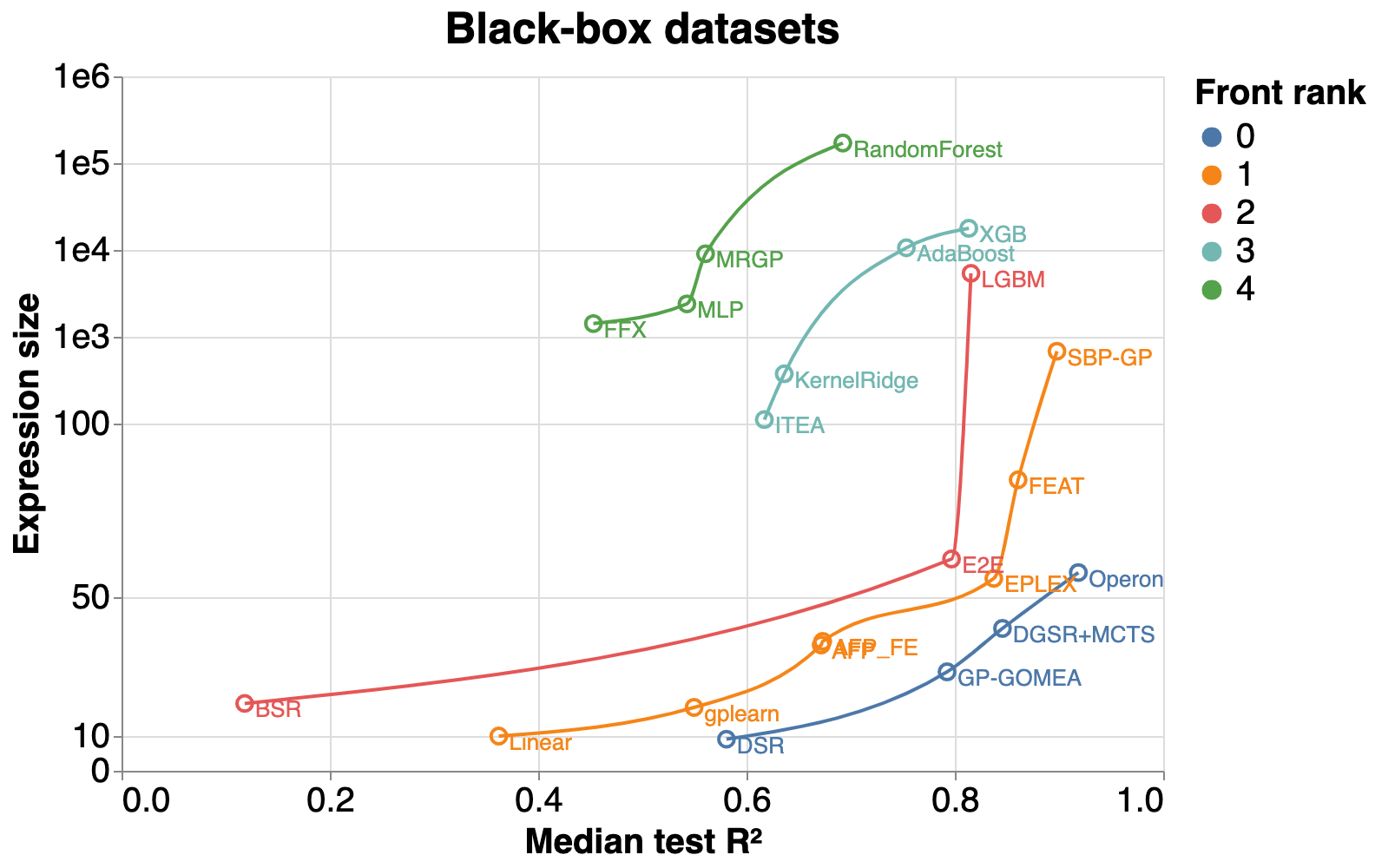}    & \includegraphics[width=0.99\columnwidth]{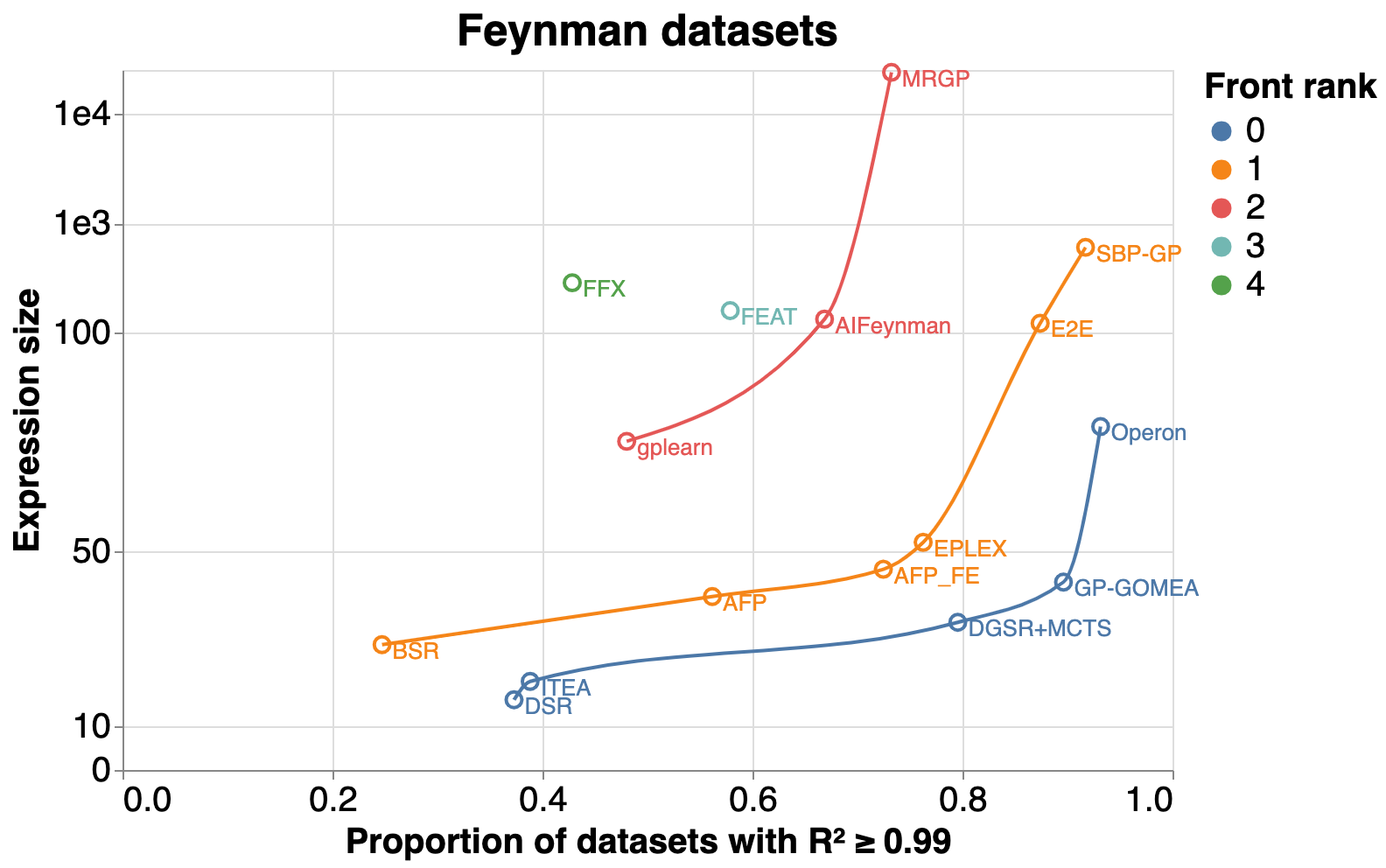} \\
\end{tabular}
\caption{
Performance on test splits of \texttt{SRBench}, respectively the median $R^2$ over black box datasets and the proportion of Feynman datasets where the $R^2$ is larger than $0.99$. To nicely visualize the trade-off between accuracy and expression size, we use a linear scale for expression size values up to $100$ then a logarithm scale. Note that AI-Feynman \cite{udrescu2020ai} was removed from the black-box plot for readability (scores $R^2=-0.6$ and expression size $744$).}
\label{fig:srbench_results}
\end{figure*}

\begin{table}[h]
    \centering
    \caption{\textbf{$\%$ of solved datasets for different mutation sizes.}}
    \begin{tabular}{c| c| c}
    \toprule
    Model & In-Domain & Out-of-domain  \\
    \midrule
    \ours{}@$1$ &$52.2$ & $26.8$ \\
    \ours{}@$10$ &$\mathbf{74.8}$ &  $\mathbf{44.0}$ \\
    \ours{}@$\infty$ & $72.4$ & $16.8$ \\
    \bottomrule
    \end{tabular}
    \label{tab:model_sizes}
\end{table}

\paragraph{How important is constant optimization?} As shown in \cite{kamienny2022end}, 
optimizing constants predicted by a NN model with an optimizer like BFGS greatly improves performance on \texttt{SRBench}. 
Similarly, we study whether including constants optimization is important for \ours{} in the context of search. 
We remark that while we use constant optimization to compute accuracy, we store the non-optimized expression in the MCTS search tree. We make this choice because optimized constants may be out-of-distribution w.r.t. the pre-trained $M_\theta$, which can lower its performance.
As shown in \Cref{tab:constant_optimisation}, optimizing expression constants improves performance substantially. However, constant optimization comes at the price of speed, especially if done after every mutation.

\begin{table}[h]
    \caption{\textbf{$\%$ of solved datasets for different constant optimization strategies.} We compare constant optimization done: never, only on the best expression of each trial and after each mutation.}
    \centering
    \resizebox{\linewidth}{!}{
    \begin{tabular}{c| c | c}
    \toprule
    Constant optimization & In-Domain & Out-of-domain  \\
    \midrule
    Never & $74.8$ &  $44.0$ \\ 
    Only best expression & $77.2$  & $59.4$ \\
    All expressions & $\mathbf{79.6}$ & $\mathbf{66.2}$   \\ 
    \bottomrule
    \end{tabular}
    }
    \label{tab:constant_optimisation}
\end{table}

\subsection{\texttt{SRBench} results}
\label{subsec:srbench_res}
We evaluate \ours{} on the regression datasets of the \texttt{SRBench} benchmark \cite{la2021contemporary},  in particular the \textit{``black-box''} datasets (no ground-truth expression is given) and the \textit{``Feynman''} datasets (conversely, the underlying physics' equation is given).
As our approach is trained on datasets with up to $10$ features, its use on higher-dimensional datasets requires feature selection.
Following \cite{kamienny2022end}, we consider only datasets with at most $10$ features so that our results are independent of the quality of a feature selection algorithm. This leads to $57$ black-box datasets and $119$ Feynman datasets.
Each dataset is split into $75\%$ training data and $25\%$ test data using sampling with a random seed (we use $3$ seeds per dataset, giving a total of $528$ datasets).
We consider all baselines provided as part of \texttt{SRBench}, which includes GP algorithms, e.g GP-GOMEA \cite{virgolin2021improving}, Operon \cite{10.1145/3377929.3398099}, ITEA \cite{10.1162/evco_a_00285}, \dgsr{} algorithms, DSR \cite{petersen2019deep} and \texttt{E2E} \cite{kamienny2022end} as well as classic machine learning regression models, e.g., multi-layer perceptron \cite{haykin1994neural} and XGBoost \cite{chen2016xgboost}.

We run \ours{} with a budget of $500,000$ evaluations (equivalently mutations) and a maximum time limit of $24$ hours. \texttt{SRBench} imposes to use at most $500 000$ evaluations per hyper-parameter configuration, and allows for six configurations, yet we provide a single configuration (resulting from \Cref{subsec:indomain}); we use the pre-trained $M_\theta$ with mutation size $10$, with $K \in  [\![8, 16]\!]$ and alternate search trials with and without constants optimization.


\begin{figure}[!ht]
\centering
\includegraphics[width=1.0\columnwidth]{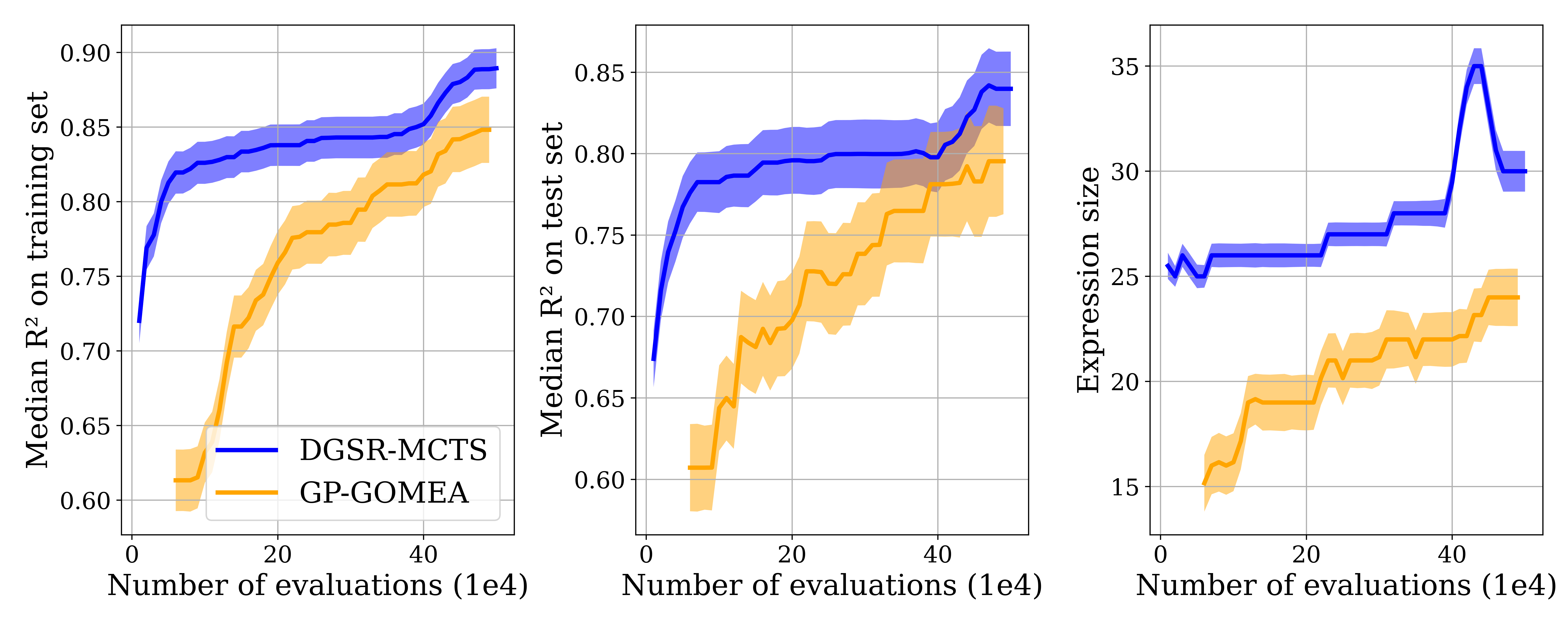}

\caption{Mean $\pm$ confidence interval performance on the black-box datasets over the number of evaluated expressions for \ours{} and its closest competitor, GP-GOMEA, on the black-box datasets.
Thanks to pre-training, \ours{} achieves high-levels of $R^2$ (and larger expressions) much more quickly than GP-GOMEA.
On the training set, \ours{} is consistently superior across the entire search process. 
On the test set and towards the end of the search, \ours{} and GP-GOMEA achieve similar results, due to a larger generalization gap for larger expressions.
}
\label{fig:srbench_learning_curves}
\end{figure}

The performance of all SR algorithms is illustrated in \cref{fig:srbench_results} along two metrics, accuracy on the test data (as measured by $R^2$) and expression size computed by counting all operators, variables and constants in an expression, after simplification by SymPy \cite{meurer2017sympy}. 
Results are aggregated by taking the average over seeds for each dataset, then the median for black-box datasets and mean for Feynman as done in \cite{la2021contemporary}. 
We visualize the trade-off between accuracy and simplicity of expressions obtained by the different algorithms; the lower and righter the better. 
We compute front ranks by Pareto-dominance. 
An algorithm Pareto-dominates another if it is not worse in all metrics and it is strictly better in at least one of them.
The definition of ranks is then recursive: an algorithm is at rank 0 if there exists no algorithm that Pareto-dominates it; for successive ranks, an algorithm is at rank $i$ if, when excluding the algorithms up to rank $i-1$, there exist no algorithm that Pareto-dominates it.
\ours{} is placed on the rank 0-front on both black-box and Feynman datasets. 
GP-GOMEA and \ours{} seem to be the best approaches for achieving simple-yet-accurate models, and interestingly switch place in their trade-off between the two metrics on the black-box and Feynman datasets. We additionally plot the performance over time on the black-box datasets against this baseline in \Cref{fig:srbench_learning_curves}.
Another interesting point is the difference between \ours{} and \texttt{E2E}; \ours{} achieve better test accuracy ($0.846$ and $0.797$ respectively) with less complex expressions ($41$ and $61$ respectively) on the black-box datasets. On $80\%$ (resp.~$87\%$ for \texttt{E2E}) of Feynman datasets, we achieve $R^2 \geq 0.99$ with expression sizes of $33$ (resp.~$121$).

\paragraph{Ablations.} Finally, we present several ablations in \Cref{tab:ablations}. Namely, we observe whether using synthetic datasets for training $C_\psi$ and fine-tuning $M_\theta$ is better than not doing so, and whether our strategy of training \ours{} simultaneously on multiple datasets is better than training iteratively, i.e., one dataset at the time.  Our findings suggest that utilizing synthetic datasets has a positive effect on the performance of our model, particularly on the Feynman datasets, which may be attributed to the similarities between the synthetic and Feynman datasets (similar expression sizes, non-noisy observations...)
On the black-box dataset (real-world scenarios), training on all datasets simultaneously appears to result in better performance than using synthetic datasets alone, likely due to the sharing of gradients across multiple datasets. Overall, our results indicate that both components play a significant role in the strong performance of \ours{}.

\begin{table}[h]
    \caption{\textbf{Ablations for different training configurations.} We report the same performance metrics used in \cref{fig:srbench_results}. 
    Respective expression sizes (not shown here) remain similar.}
    \centering
    \resizebox{\linewidth}{!}{
    \begin{tabular}{c | c| c | c}
    \toprule
   \makecell{Use synthetic \\ datasets}   & \makecell{Simultaneous \\ training} & Black-box & Feynman  \\
    \midrule
   no & no & $0.801$ & $0.655$ \\
   yes & no & $0.812$ & $0.748$ \\
   no & yes & $0.823$ & $0.689$ \\
   yes & yes & $\mathbf{0.846}$  & $\mathbf{0.796}$\\
    \bottomrule
    \end{tabular}
    }
    \label{tab:ablations}
\end{table}

\section{Limitations}
Our approach is subject to the known limitations of the Transformer models as in \cite{kamienny2022end}. For instance, learning on large context lengths is challenging and necessitates significant GPU memory resources. This limitation could be circumvented by the use of Transformers specifically designed for large inputs, such as LongFormer \cite{beltagy2020longformer}. 
The main source of latency in our algorithm comes from sampling mutations (i.e. forwarding the Transformer model), an aspect shared by other DGSR methods that use large pre-trained transformers. While GP approaches can run on CPU only, GPUs can be useful to batch computations (to generate mutations in our case) for DGSR methods.

\section{Conclusion}
In this work, we introduced a competitive SR algorithm that address the limits of \dgsr{} by incorporating search as a tool to further improve performance of inductive \dgsr{} approaches. We positioned our method within a unifying view of SR algorithms that, we hope, will shed light on inner workings of different class of algorithms. 
We showed \ours{} performs on par with state-of-the-art SR algorithms such as GP-GOMEA in terms of accuracy-complexity trade-off, while being more efficient in terms of number of expression evaluations. 

Future work may concern the extension of the proposed approach in a meta-learning framework \cite{schmidhuber:1987:srl,thrun2012learning,finn2017model}, where pre-training is performed, either via self-supervised or reinforcement learning, with the objective to reduce the required search budget on a broad family of target datasets.


\section{Acknowledgments}
We acknowledge Fabricio Olivetti de França from Universidade Federal do ABC for providing the code to produce \cref{fig:srbench_results}.

\clearpage
\bibliographystyle{icml2023}
\bibliography{ref}

\clearpage
\appendix

\section{Data generation}
\label{ap:data_generation}

\paragraph{Ground-truth expressions.}
To generate a large synthetic dataset with examples $(\mathcal{D}, f^*)$, we first sample $N$ observations/features $\mathbf{X} \in \mathrm{R}^{N\times D}$ where $D$ is uniformly sampled between $1$ and $10$  with a mixture of Gaussians as in \cite{kamienny2022end}, then consider sampling: a) an empty unary-binary tree from \cite{lample2019deep} generator with between $5$ and $25$ internal nodes, b) assign a random operator on nodes and either a variable $\{x_d\}_{d \leq D}$ or float constant drawn from a normal distribution on leaves. We then simplify the ground-truth expression with SymPy.

The only difference with \cite{kamienny2022end} lies in the fact that the generated expressions are much smaller by not enforcing all variables to appear sampled expressions, therefore letting the model learn the ability to do feature selection, as well as to having much less constants (they apply linear transformations with probability $0.5$  on all nodes/leaves), therefore providing more interpretable expressions (model size is divided by $2$).


\paragraph{Example mutations.} From a ground-truth expression $f^*$, we generate a sequence of example mutations that go from the empty expression to $f^*$ by iteratively removing parts of the expression tree. To do so, at any given step, we randomly pick an internal node such that the size of the subtree argument $B$ is large enough, and apply the backward mutation, i.e. adding an edge between the parent node and the remaining child node $A$. When the remaining expression is too small, the mutation's operator becomes  $0 \xrightarrow[]{} B $ where $B$ is the remaining expression.

\section{Data representation}
\label{ap:data_representation}
As done in most \cite{kamienny2022end}, floats are represented in base 10 foating-point notation, rounded to four significant digits, and encoded as sequences of 3 tokens: their sign, mantissa (between $0$ and $9999$), and exponent (from \texttt{E-100} to \texttt{E100}). Expressions are represented by their Polish notations, i.e. the breadth-first search walk, with numerical constants are represented as explained above. A dataset is represented by the concatenation of all tokenized $(x_i, y_i)$ pairs where vectors representation is just the flattenized tokens of each dimension value.  The combination of both the expression and dataset yields the representation of states by concatenating both representations and using special separators between the expression $f$ and dataset $\mathcal{D}$. Actions are represented by the concatenation (with special separators) of i) the node index (integer in base $10$) on which to apply the mutation, ii) the operator token, iii) the tokenized expression $B$ if the operator is binary.

\section{Model details}
\label{ap:model_details}

Since the number of tokens transformers can use as context is limited by memory considerations and possible learnable long-range dependencies, we restrict to $100$ the number of input data points used as input to $M$, the subset being sampled at each expansion. We also train our model on datasets with at most $10$ variables, as in \cite{kamienny2022end}.

\section{Exceptions detected ablations} 
\label{ap:exceptions} 

Earlier \dgsr{} work has showed that Transformer models \cite{vaswani2017attention} were able to learn almost perfect semantics of symbolic expressions, resulting in $99\%$ of valid expressions in \cite{kamienny2022end}. 
In this work, we noticed that our policy model was also able to manipulate the expression structure quite well as around $90\%$ of sampled mutations resulted in valid expressions. Similarly we noticed that malformed mutations, e.g., invalid index on which expression node to apply an operation, argument $B$ that cannot be parsed or or absent $B$ when operator is binary, represent less than $1\%$ of errors.

As mentioned in \cref{subsec:method_mutation}, we constrain our model to discard expressions that are too complex (more than $60$ operators/variables/constants) or have nested complicated operators (e.g. $\cos(\cos(X))$, $\log(\log(X))$), in order to promote simpler expressions as explained, resulting in a great trade-off between accuracy and complexity as shown in \cref{subsec:srbench_res}. Note that it would be possible to enforce a greater trade-off between accuracy and complexity, e.g. using strategies mentioned in \cite{la2021contemporary}. This results in $9\%$ of expressions being discarded because these constraints are violated.

\section{Search details}
\label{ap:search_details}

\subsection{Definition of satisfactory expressions}
We concluded from preliminary experiments that considering $f^*$ to be satisfactory if $R^2\geq0.99$ performed best as it provided high-quality samples while dramatically reducing search times compared to perfect fitting. Systematically estimating what accuracy can be achieved with a given complexity is not possible without, e.g., resorting to another algorithm that operates under the same complexity constraints and can act as an oracle. 
We also tried estimating accuracy on validation set of points by running XGBoost on a train set, however for \texttt{SRBench} datasets, accuracy can greatly vary according to the way the dataset is split.

\subsection{Constant optimization}
\label{ap:constant_optimization}
We use Broyden–Fletcher–Goldfarb–Shanno algorithm (BFGS) with batch size $256$, early stopping if accuracy does not improve after $10$ iterations and a timeout of $1$ second. 

\subsection{Search hyper-parameters}
In this work, we employ a distributed learning architecture, similar to that proposed in \cite{lample2022hypertree}. 
Since the optimal hyper-parameters of search are not necessarily the same for all datasets, the controller samples these hyper-parameters from pre-defined ranges for each different search trial:

The proposed model depends on many of hyper-parameters, specifically those pertaining to the decoding of the mutation model and the search process. Determining the optimal values for these hyper-parameters poses a significant challenge in practice for several reasons. Firstly, the model is in a state of continual evolution, and thus the optimal hyper-parameter values may also change over time. For instance, if the model exhibits an excessive level of confidence in its predictions, it may be necessary to increase the decoding temperature to promote diversity in mutations. Secondly, the optimal values of the hyper-parameters may be specific to the dataset under consideration. Finally, the sheer number of hyper-parameters to be tuned, coupled with the high computational cost of each experiment, renders the task of determining optimal values infeasible. To circumvent these issues, rather than fixing the hyper-parameters to a specific value, they are sampled from predefined ranges at the beginning of each search trial. The specific decoding parameters and the distribution utilized are as follows:
\begin{itemize}
\item Number of samples $K$ per expansion. Distribution: uniform on range [8,16].
\item Temperature used for decoding. Distribution: uniform on range [0.5, 1.0].
\item Length penalty: length penalty used for decoding. Distribution: uniform on range [0, 1.2].
\item Depth penalty: an exponential value decay during the backup-phase, decaying with depth
to favor breadth or depth. Distribution: uniform on discrete values [0.8, 0.9, 0.95, 1].
\item Exploration: the exploration constant $p_{uct}$. 1

\end{itemize}



\end{document}